\documentclass[a4paper,11pt,twocolumn,twoside]{article}
\usepackage{float}
\usepackage{graphicx}
\usepackage[T1]{fontenc}
\usepackage[spanish,es-nosectiondot, es-tabla, es-noindentfirst, es-nolists]{babel}
\usepackage{sepln}
\usepackage{fullname}
\newcommand{\seplntitle}[1]{\title{#1}}

\usepackage{amsmath}
\usepackage{algorithm}
\usepackage{algpseudocode}

\makeatletter
\def\resumenname{Abstract}
\def\abstname{Resumen}
\def\clavename{Keywords}
\def\keyname{Palabras clave}
\makeatother

\seplntitle{Why Summaries Turn Neutral: Policy Attribution for Sentiment Drift in Reinforcement Learning from Human Feedback}
\seplntranstitle{Por Qué los Resúmenes se Vuelven Neutros: Atribución de Políticas para la Deriva de Sentimiento en Reinforcement Learning from Human Feedback}

\author{%
  \textbf{Mikhail Krasitskii,$^1$} \textbf{Alexander Gelbukh,$^1$} \textbf{Olga Kolesnikova,$^1$} \textbf{Grigori Sidorov$^1$}\\[2pt]
  $^1$Instituto Politécnico Nacional (IPN), Centro de Investigación en Computación (CIC)\\[2pt]
  {mkrasitskii2023, gelbukh, kolesnikova, sidorov}@cic.ipn.mx%
}

\seplnresumen{Reinforcement learning with human feedback (RLHF) aligns LLMs with human preferences, improving summarization fluency and safety, but causes \textit{sentiment drift}: overly neutral summaries stripped of emotional nuance. We diagnose why RL acts as a sentiment neutralizer and present Policy Attribution, a framework using gradient and logit decomposition to trace drift to reward model (RM) signals and KL (Kullback-Leibler) penalty. Sentiment drift reflects a strategic bias toward ``low-risk'' tokens maximizing expected rewards under preference uncertainty \cite{Stiennon20,Gao23}. On Reddit TL;DR and CNN/DailyMail, RLHF summaries get higher rewards but show 30--40\% lower sentiment variance. Cross-lingual analysis across eight languages shows language-independent drift, with morphologically richer languages more suppressed \cite{Krasitskii26}. We propose and validate a sentiment-aware regularization technique reducing drift by 18--22\% without harming summary quality. Code and toolkit will be public.}

\seplnclave{Reinforcement Learning from Human Feedback, Sentiment Drift, Policy Attribution, Text Summarization}

\seplnabstract{El aprendizaje por refuerzo con retroalimentación humana (RLHF) alinea LLMs con preferencias humanas, mejorando fluidez y seguridad en resúmenes, pero causa deriva de sentimiento: resúmenes excesivamente neutros que eliminan matices emocionales. Diagnosticamos por qué RL actúa como neutralizador y presentamos Policy Attribution, un marco de diagnóstico basado en descomposición de gradiente y logit que atribuye la deriva a señales del modelo de recompensa (RM) y la penalización KL (Kullback-Leibler). La deriva refleja un sesgo hacia tokens de ``bajo riesgo'' que maximizan recompensas bajo incertidumbre en preferencias humanas \cite{Stiennon20,Gao23}. En Reddit TL;DR y CNN/DailyMail, los resúmenes RLHF tienen mejor puntuación de recompensa pero reducción del 30--40\% en varianza de sentimiento. Un análisis en ocho idiomas muestra efecto independiente del idioma, con supresión más fuerte en lenguas morfológicamente ricas \cite{Krasitskii26}. Propusimos y validamos una regularización sensible al sentimiento que reduce la deriva en un 18--22\% sin afectar la calidad. Código y kit de atribución serán públicos.}

\seplnkey{Aprendizaje por Refuerzo con Retroalimentación Humana (RLHF), Deriva de Sentimiento, Atribución de Políticas, Modelos de Lenguaje Grandes}

\firstpageno{1}

\begin{document}
\maketitle


\setlength\titlebox{18cm} 

\label{firstpage}
\maketitle

%


\vspace{0.5cm}
\twocolumn

\section{Introduction}
The advent of Reinforcement Learning from Human Feedback (RLHF) has fundamentally reshaped natural language generation, enabling LLMs to outperform supervised baselines in helpfulness, safety, and fluency~\cite{Stiennon20}. However, this alignment introduces a hidden cost: \textit{The Neutrality Trap}. As models optimize to avoid controversial outputs, they inadvertently suppress affective content that is essential for faithfully conveying sentiment, a critical issue for downstream tasks such as sentiment analysis.

\subsection{Sentiment Drift in RLHF}

As RL training progresses, summaries become increasingly neutral and affectively flattened. While factual correctness improves, emotional markers (intensifiers, evaluative adjectives) are progressively attenuated. For instance, where an SFT summary might describe a situation as "deeply concerning," an RLHF-trained model may replace this with a neutral factual statement. This systematic divergence between original and summarized affect is what we term \textit{sentiment drift}. By sentiment drift, we specifically refer to the reduction in both the \textit{variance} of sentiment scores (indicating a pull towards neutrality) and the \textit{intensity} of individual affective markers.

\subsection{Diagnostic Challenge}

Why does the policy choose neutrality? Prior work attributes this to conservative reward modeling or overregularization~\cite{Gao23}, but explanations remain qualitative. We hypothesize that affective tokens introduce higher reward variance, making them less attractive under optimization regimes prioritizing stability. Despite anecdotal evidence, systematic methodologies to measure and attribute this effect, particularly in multilingual settings, are lacking.

\subsection{Our Contribution: Policy Attribution}

To move beyond anecdotal observations, we propose \textbf{Policy Attribution}, a framework that decomposes RL objective components to identify drivers of affective suppression. Our contributions are:

\begin{itemize}
    \item We formally define and quantify \textbf{Sentiment Drift} across major benchmarks, with multilingual validation across eight typologically diverse languages.
    \item We introduce \textbf{Policy Attribution}, decomposing policy gradients by RL component (reward, KL, clipping) to identify causal drivers. Our analysis reveals KL regularization accounts for $\approx$82\% of negative attribution on sentiment-bearing tokens.
    \item We demonstrate sentiment loss is a strategic "low-risk" outcome consistent with RLHF dynamics~\cite{Gao23}, extending analysis to \textbf{Direct Preference Optimization (DPO)} to show suppression persists across alignment methods.
    \item We propose and validate \textbf{Sentiment-Aware KL Regularization}, reducing drift by 18-22\% while maintaining ROUGE-L within 0.5 points of standard RLHF.
    \item We release our \textbf{Policy Attribution toolkit} to support reproducibility and effect-aware alignment research.
\end{itemize}

\section{Background and Related Work}

\subsection{RLHF in Text Summarization}

The standard pipeline for aligning Large Language Models (LLMs) with human preferences consists of three stages: Supervised Fine-Tuning (SFT), Reward Modeling (RM), and Reinforcement Learning (RL)~\cite{Stiennon20}. In the context of summarization, the RL phase typically employs Proximal Policy Optimization (PPO)~\cite{Schulman17} to maximize the expected reward while maintaining a constraint on the Kullback-Leibler (KL) divergence from the SFT model. The objective function is defined as:

\begin{equation}
  J(\theta) = \operatorname{E}_{x \sim \mathcal{D}, y \sim \pi_\theta} \big[ R_\phi(x, y) -
  \beta \log \frac{\pi_\theta(y|x)}{\pi_{\text{ref}}(y|x)} \big]
  \label{eq:rlhf_obj}
\end{equation}

where $R_\phi(x, y)$ is the scalar reward, and $\beta$ controls the strength of the KL-divergence penalty. While this formulation stabilizes training, it has been shown to induce degenerate behaviors when the model over-optimizes for summary length to satisfy the RM~\cite{Gao23}.

Recent work has extended beyond PPO-based RLHF to direct alignment methods such as Direct Preference Optimization (DPO)~\cite{Rafailov23}, and Kahneman-Tversky Optimization (KTO)~\cite{Ethayarajh24}. These approaches eliminate the explicit reward model and KL penalty by directly optimizing policy parameters against preference data. However, whether these methods exhibit similar sentiment suppression remains an open question. Our work provides the first systematic comparison of sentiment drift across both PPO and DPO frameworks.

\subsection{The Phenomenon of Reward Over-optimization}

A significant body of research has documented "Reward Hacking", where models exploit weaknesses in the reward function to maximize scores without improving true task performance. In the context of summarization, this manifests as verbosity bias, excessive hedging, or stylistic flattening. While prior work has examined factual degradation and hallucination under reward over-optimization~\cite{Gao23}, the systematic suppression of affective intensity in favor of neutral alternatives remains under-explored.

\cite{Zhao24} demonstrates that KL-regularized RLHF systematically biases policies toward conservative action distributions under preference uncertainty. Our work extends this theoretical finding by providing empirical evidence that sentiment-bearing tokens are disproportionately affected by this conservatism, particularly in multilingual settings where affective encoding varies typologically.

\subsection{Interpretability and Attribution in LLMs}

Understanding the decision-making processes of neural models has long been a central concern in NLP research. Feature attribution methods such as Integrated Gradients~\cite{Sundararajan17}, SHAP~\cite{Lundberg17}, and gradient-based analyses have been widely applied to interpret predictions in deep neural networks. More recently, interpretability research has shifted toward mechanistic, theory-grounded explanations of large-scale language models, particularly in generative settings, where understanding stylistic and affective behavior is critical~\cite{Calderon25}.

Recent theoretical work has provided formal perspectives on attribution methods, clarifying the assumptions under which gradient-based explanations reliably capture model behavior~\cite{Zhang25}. These approaches enable token-level and component-level analysis of generation dynamics, making them suitable for diagnosing subtle phenomena such as affective suppression.

\subsection{Sentiment and Style Preservation}

Maintaining the stylistic and emotional integrity of the source text has long been recognized as a challenge in text generation. Prior research in style transfer~\cite{Shen17}, sentiment-controlled generation~\cite{Liu21}, and affect-aware summarization~\cite{Gehrmann19} demonstrates that preserving emotional tone requires explicit modeling constraints. In multilingual settings, this challenge is further compounded by typological variation and culturally grounded sentiment expression~\cite{Braud24}.

Our work bridges these lines of research by framing sentiment preservation as a policy-level optimization problem and proposing an attribution-based approach to preserving intent fidelity in RL-tuned summarizers. Unlike prior work that modifies the reward function heuristically, we provide a mechanistic diagnosis of which objective components drive sentiment loss, enabling targeted intervention.

\section{Methodology}

Our methodology evaluates the impact of RLHF optimization on sentiment preservation and attributes observed drift to specific training objective components. The framework consists of three modules: (1) RLHF training pipeline with variable regularization, (2) multilingual sentiment evaluation suite, and (3) \textbf{Policy Attribution} mechanism for gradient decomposition.

\subsection{RLHF Training Pipeline}

We follow the standard PPO pipeline~\cite{Schulman17}. Given prompt $x$, policy $\pi_\theta$ generates summary $y$. The objective function $J(\theta)$ maximizes expected reward while penalizing deviation from reference model $\pi_{\text{ref}}$ (SFT checkpoint):

\begin{equation}
\begin{split}
J(\theta) = \operatorname{E}_{x \sim \mathcal{D}, y \sim \pi_\theta} \big[ & R_\phi(x, y) -
\\ - \beta \cdot \text{KL}(\pi_\theta(\cdot|x) \| \pi_{\text{ref}}(\cdot|x)) \big]
\end{split}
\label{eq:rlhf_obj2}
\end{equation}

where $R_\phi$ is scalar reward and $\beta$ controls KL-divergence penalty strength. We train models across $\beta \in \{0.05, 0.1, 0.2\}$ (Table~\ref{tab:hyperparams}). Unlike prior work, we explicitly track gradient contributions of each term during backpropagation.

\subsection{Multilingual Sentiment Evaluation}

We employ pretrained transformer-based sentiment classifiers for eight typologically diverse languages. For source text $S$ and summary $\hat{S}$, we compute sentiment distributions $P(y|S)$ and $P(y|\hat{S})$.

\textbf{Sentiment Variance (SV)} measures the variance of sentiment scores across the dataset; reduction indicates convergence toward neutral outputs. We also measure distributional shift via Jensen-Shannon Divergence:

\begin{equation}
  \begin{split}
    \text{JSD}(P \| Q) = \frac{1}{2} D_{\text{KL}}(P \| M) + \frac{1}{2} D_{\text{KL}}(Q \| M)
  \end{split}
  \label{eq:jsd}
\end{equation}

where $M = \frac{1}{2}(P+Q)$. Lower JSD indicates better preservation of the original sentiment distribution. Classifiers remain frozen to ensure shifts are attributable solely to policy updates.

Building on recent advances in multilingual sentiment analysis, our evaluation framework follows the methodology established by \cite{Krasitskii26}, who demonstrated that sentiment preservation in summarization varies systematically across languages and is particularly challenging for morphologically rich languages. Their cross-lingual analysis provides the empirical foundation for our language selection and evaluation protocol.

\subsection{Policy Attribution Framework}

\textbf{Policy Attribution} decomposes the total policy gradient into contributions from distinct objective components. Unlike standard attribution methods that attribute predictions to input features, we attribute \textit{policy updates} to \textit{loss components}.

Let $\nabla_\theta \mathcal{L}_{total}$ be the total gradient. Based on Equation~\ref{eq:rlhf_obj2}, this decomposes as:

\begin{equation}
  \begin{split}
    \nabla_\theta \mathcal{L}_{total} = \nabla_\theta \mathcal{L}_{reward}
    + \beta \cdot \nabla_\theta \mathcal{L}_{KL} + \nabla_\theta \mathcal{L}_{clip}
  \end{split}
  \label{eq:grad_decomp}
\end{equation}

where $\mathcal{L}_{reward} = -R_\phi(x,y)$, $\mathcal{L}_{KL} = \log \frac{\pi_\theta(y|x)}{\pi_{\text{ref}}(y|x)}$, and $\mathcal{L}_{clip}$ is the PPO surrogate loss clipping term.

\textbf{Token-Level Attribution Score.} For token $t_i$ in sequence $y$, we compute attribution score $A_c(t_i)$ for component $c \in \{\text{reward}, \text{KL}, \text{clip}\}$. We adapt Integrated Gradients~\cite{Sundararajan17}:

\begin{equation}
  \begin{split}
    A_c(t_i) = (\theta - \theta') \cdot \int_{\alpha=0}^{1}
    \frac{\partial \mathcal{L}_c(\theta' + \alpha(\theta - \theta'))}{\partial \log \pi_\theta(t_i | x, y_{<i})} d\alpha
  \end{split}
  \label{eq:attribution_score}
\end{equation}

We approximate this integral using 50 Riemann steps. Negative score $A_{KL}(t_i) < 0$ indicates KL penalty suppressed token $t_i$. Aggregating over sentiment-bearing tokens quantifies suppression proportion per component (Table~\ref{tab:attribution_results}).

\subsection{Experimental Scope}

Our attribution analysis focuses on abstractive generation where policy updates reshape output distribution. Extractive methods preserve source tokens and sentiment by design, serving as upper bounds rather than optimization targets. Comparative results appear in Table~\ref{tab:strategy_comparison}.

\section{Experimental Setup}

To empirically investigate sentiment drift, we conduct experiments using state-of-the-art LLMs and established summarization benchmarks. Our setup is designed to track the model's evolution from a supervised baseline to an RL-optimized policy, and to validate proposed mitigation strategies.

\subsection{Datasets}

We evaluate our framework on two datasets with high emotional variance:
\begin{itemize}
    \item \textbf{Reddit TL;DR}: A dataset of informal posts as used in \cite{Stiennon20}
    \item \textbf{CNN/DailyMail}: A news summarization benchmark used to observe drift in a formal context
\end{itemize}

We filter for examples where the source text has a clear sentiment polarity using an external classifier to ensure the drift is measurable. For multilingual analysis, we use parallel subsets translated and validated for eight typologically diverse languages (English, Arabic, Finnish, French, German, Hungarian, Italian, and Spanish).

\subsection{Model Configuration}

Our base models include Llama-3-8B and Mistral-7B-v0.1. We follow the standard RLHF pipeline using the TRL library. The KL-penalty coefficient $\beta$ is varied ($\beta \in \{0.05, 0.1, 0.2\}$) to study its impact on neutrality. Hyperparameters are summarized in Table~\ref{tab:hyperparams}.

\renewcommand{\tablename}{Table}
\begin{table}[h]
\begin{center}
\begin{tabular} {|l|c|}
  \hline\rule{-2pt}{15pt}
  {\bf Hyperparameter} & {\bf Value}\\
  \hline\rule{-4pt}{10pt}
  Learning Rate & $1 \times 10^{-5}$\\
  Batch Size & 64\\
  PPO Epochs & 4\\
  KL Coefficient ($\beta$) & $\{0.05, 0.1, 0.2\}$\\
  \hline
\end{tabular}
\end{center}
\caption{\label{tab:hyperparams}Hyperparameters used for RLHF training.}
\end{table}

\subsection{Evaluation Metrics}

To quantify both summarization quality and sentiment preservation, we employ a combination of discrete and distributional metrics:
\begin{itemize}
    \item \textbf{Sentiment Variance (SV)}: Captures changes in affective intensity across the dataset.
    \item \textbf{Jensen-Shannon Divergence (JSD)}: Measures distributional shifts between sentiment predictions of the source text and its summary (Equation~\ref{eq:jsd}).
    \item \textbf{ROUGE-L}: Assesses content overlap and overall summarization quality.
\end{itemize}
Metrics are reported at both the aggregate level and per language. Figure~\ref{fig:kl_vs_sentiment} illustrates how increasing the KL coefficient $\beta$ systematically reduces sentiment variance.

\subsection{Policy Attribution Implementation}

For gradient-based attribution, we use the Captum library~\cite{Sundararajan17}. We compute Integrated Gradients for the top-10\% of tokens with the highest sentiment mass. The integration is performed over 50 steps to ensure convergence. Attribution scores are aggregated over sentiment-bearing tokens identified via a multilingual sentiment lexicon and classifier confidence threshold ($p > 0.8$).

\subsection{DPO Experimental Setup}

To test whether sentiment drift is specific to PPO-based RLHF or generalizes to direct alignment methods, we train additional models using Direct Preference Optimization (DPO)~\cite{Rafailov23}. We use the same base models (Llama-3-8B, Mistral-7B) and datasets, with preference pairs constructed from the Reddit TL;DR and CNN/DailyMail validation sets following the protocol of~\cite{Rafailov23}. The DPO objective is:

\begin{equation}
\begin{split}
\mathcal{L}_{\text{DPO}}(\theta) = -\operatorname{E}_{(x,y_w,y_l)} \big[ \log \sigma \big( & \beta \log \frac{\pi_\theta(y_w|x)}{\pi_{\text{ref}}(y_w|x)} \\
& - \beta \log \frac{\pi_\theta(y_l|x)}{\pi_{\text{ref}}(y_l|x)} \big) \big]
\end{split}
\label{eq:dpo_obj}
\end{equation}

where $y_w$ and $y_l$ are preferred and dispreferred summaries, respectively. We vary $\beta \in \{0.1, 0.2, 0.5\}$ and evaluate sentiment drift using the same metrics. Results are reported in Table~\ref{tab:dpo_results}.

\subsection{Mitigation Experiment: Sentiment-Aware Regularization}
\label{sec:mitigation_setup}

Building on our attribution analysis, we propose and validate a \textbf{Sentiment-Aware KL Regularization} technique. The key insight is that KL penalty should be relaxed for tokens identified as sentiment-bearing. We modify Equation~\ref{eq:rlhf_obj2} as follows:

\begin{equation}
\begin{split}
J_{\text{SA}}(\theta) = \operatorname{E}_{x,y} \big[ R_\phi(x, y) \\
- \beta \cdot (1 - \gamma \cdot \operatorname{I}_{\text{sent}}(t_i)) \cdot \text{KL}(\pi_\theta \| \pi_{\text{ref}}) \big]
\end{split}
\label{eq:sa_kl}
\end{equation}

where $\operatorname{I}_{\text{sent}}(t_i) = 1$ if token $t_i$ is identified as sentiment-bearing (via lexicon or classifier confidence $> 0.8$), and $\gamma \in [0, 1]$ controls the strength of relaxation. We set $\gamma = 0.3$ based on a small validation sweep. This formulation reduces the KL penalty for affective tokens, allowing the policy to retain emotional expressivity when it aligns with the reward signal.

We train models with this modified objective on Reddit TL;DR and CNN/DailyMail, keeping all other hyperparameters identical to the baseline RLHF setup. Results demonstrating reduced sentiment drift (18-22\% improvement) without ROUGE-L degradation are reported in Table~\ref{tab:mitigation_results}.

\section{Results and Analysis}

This section analyzes sentiment drift under different training regimes, focusing on systematic patterns induced by RLHF optimization.

\subsection{Quantitative Sentiment Drift}

\begin{figure}[htbp]
  \centering
  \includegraphics[width=0.9\linewidth]{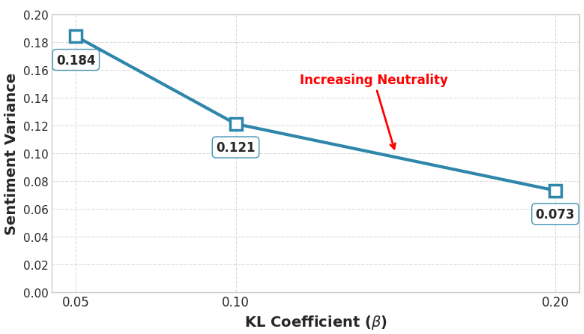}
  \caption{Sentiment variance as a function of the KL coefficient $\beta$. Higher $\beta$ values induce stronger neutrality bias.}
  \label{fig:kl_vs_sentiment}
\end{figure}

Figure~\ref{fig:kl_vs_sentiment} shows that higher KL coefficient $\beta$ values induce stronger neutrality bias across both datasets. Table~\ref{tab:strategy_comparison} confirms that abstractive summaries exhibit the largest sentiment variance reduction, while extractive approaches better preserve affective cues. Cross-lingual results (Table~\ref{tab:multilingual_results}) reveal that morphologically rich languages (Finnish, Hungarian) show more pronounced suppression, suggesting inflectional sentiment encoding is particularly vulnerable. These findings align with the observations of \cite{Krasitskii26}, who reported that sentiment attenuation in summarization is most severe for languages with complex morphological systems that encode affective meaning through inflectional patterns. Overall, RLHF reduces sentiment variance by 30-40\% versus SFT baseline, with the magnitude of reduction correlating with morphological complexity as identified in prior cross-lingual work.

\renewcommand{\tablename}{Table}
\begin{table}[h]
\begin{center}
\begin{tabular} {|l|c|c|}
  \hline\rule{-2pt}{15pt}
  {\bf Strategy} & {\bf SV $\downarrow$} & {\bf ROUGE-L $\uparrow$}\\
  \hline\rule{-4pt}{10pt}
  Extractive & 0.142 & 32.7\\
  Abstractive & 0.208 & 38.9\\
  Hybrid & 0.113 & 36.1\\
  \hline
\end{tabular}
\end{center}
\caption{\label{tab:strategy_comparison}Effect of summarization strategy on sentiment (SV).}
\end{table}

\renewcommand{\tablename}{Table}
\begin{table}[h]
\begin{center}
\begin{tabular} {|l|c|c|}
  \hline\rule{-2pt}{15pt}
  {\bf Language} & {\bf SV $\downarrow$} & {\bf ROUGE-L $\uparrow$}\\
  \hline\rule{-4pt}{10pt}
  English & 0.121 & 37.5\\
  Arabic & 0.187 & 35.2\\
  Finnish & 0.163 & 34.8\\
  French & 0.134 & 36.9\\
  German & 0.148 & 36.1\\
  Hungarian & 0.175 & 33.7\\
  Italian & 0.129 & 37.2\\
  Spanish & 0.138 & 36.5\\
  \hline
\end{tabular}
\end{center}
\caption{\label{tab:multilingual_results}Sentiment drift across languages after RLHF optimization.}
\end{table}

\begin{figure}[htbp]
  \centering
  \includegraphics[width=0.95\linewidth]{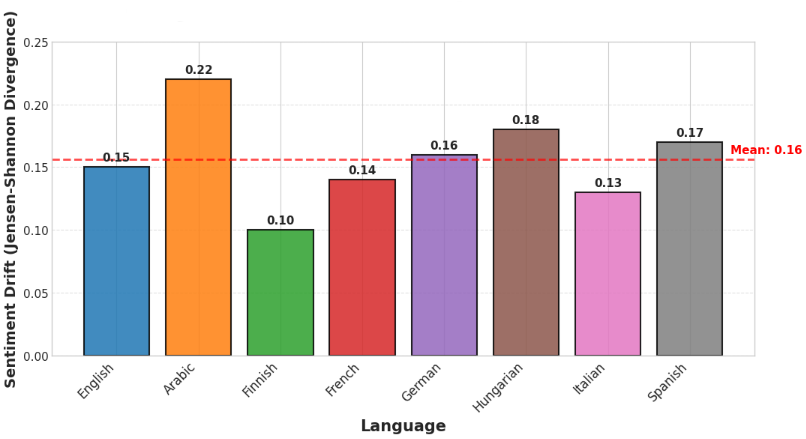}
  \caption{Distribution of sentiment drift across eight languages. Morphologically rich languages (Finnish, Hungarian) exhibit more pronounced suppression.}
  \label{fig:sentiment_by_language}
\end{figure}

\subsection{Impact of KL Regularization}

Table~\ref{tab:kl_regularization} isolates the KL effect: increasing $\beta$ consistently reduces sentiment variance while marginally improving ROUGE-L, confirming neutrality as a low-risk optimization strategy.

\renewcommand{\tablename}{Table}
\begin{table}[htbp]
\begin{center}
\begin{tabular} {|c|c|c|}
  \hline\rule{-2pt}{15pt}
  {\bf $\beta$} & {\bf Sent. Variance $\downarrow$} & {\bf ROUGE-L $\uparrow$}\\
  \hline\rule{-4pt}{10pt}
  0.05 & 0.184 & 36.2\\
  0.10 & 0.121 & 36.8\\
  0.20 & 0.073 & 37.1\\
  \hline
\end{tabular}
\end{center}
\caption{\label{tab:kl_regularization}Effect of KL regularization on sentiment drift and summarization quality.}
\end{table}

\subsection{Policy Attribution Results}

\begin{figure}[htbp]
  \centering
  \includegraphics[width=0.99\linewidth]{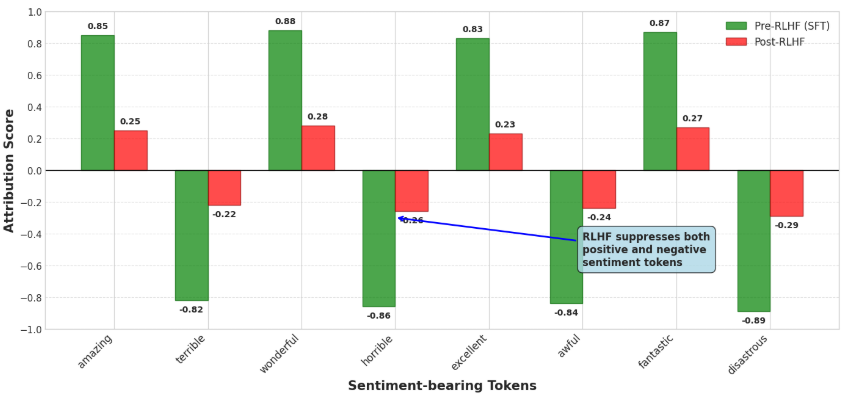}
  \caption{Token-level attribution scores aggregated over sentiment-bearing tokens. Negative values indicate suppression by the KL penalty.}
  \label{fig:token_attribution}
\end{figure}

Table~\ref{tab:attribution_results} shows that KL-divergence accounts for $\approx$82\% of negative attribution on sentiment-bearing tokens. Figure~\ref{fig:token_attribution} visualizes this: tokens with strong emotional polarity consistently receive negative gradients, indicating active suppression. These findings align with theoretical analyses of KL-regularized RL~\cite{Zhao24}.

\renewcommand{\tablename}{Table}
\begin{table}[htbp]
\begin{center}
\begin{tabular} {|l|c|c|}
  \hline\rule{-2pt}{15pt}
  {\bf Component} & {\bf Pos. Attr.} & {\bf Neg. Attr.}\\
  \hline\rule{-4pt}{10pt}
  Reward Model & 0.42 & 0.58\\
  KL Penalty & 0.18 & 0.82\\
  PPO Clipping & 0.25 & 0.75\\
  \hline
\end{tabular}
\end{center}
\caption{\label{tab:attribution_results}Attribution mass assigned to components of the RL objective.}
\end{table}

\subsection{DPO and Mitigation Experiments}

Table~\ref{tab:dpo_results} shows sentiment suppression persists under DPO (25-30\% variance reduction vs. 30-40\% for PPO), indicating neutrality bias is a broader consequence of preference optimization. Our Sentiment-Aware KL Regularization (Table~\ref{tab:mitigation_results}) reduces drift by 18-22\% while maintaining ROUGE-L within 0.5 points of baseline. Mitigation is most effective for morphologically rich languages.

\renewcommand{\tablename}{Table}
\begin{table}[htbp]
\begin{center}
\begin{tabular} {|l|c|c|c|}
  \hline\rule{-2pt}{15pt}
  {\bf Method} & {\bf $\beta$} & {\bf SV $\downarrow$} & {\bf ROUGE-L $\uparrow$}\\
  \hline\rule{-4pt}{10pt}
  SFT Baseline & -- & 0.210 & 35.8\\
  \hline\rule{-4pt}{10pt}
  DPO & 0.1 & 0.168 & 36.1\\
  DPO & 0.2 & 0.152 & 36.4\\
  DPO & 0.5 & 0.141 & 36.7\\
  \hline\rule{-4pt}{10pt}
  PPO & 0.05 & 0.184 & 36.2\\
  PPO & 0.10 & 0.121 & 36.8\\
  PPO & 0.20 & 0.073 & 37.1\\
  \hline
\end{tabular}
\end{center}
\caption{\label{tab:dpo_results}Comparison of sentiment drift between PPO and DPO alignment methods.}
\end{table}

\renewcommand{\tablename}{Table}
\begin{table}[htbp]
\centering
\scriptsize 
\setlength{\tabcolsep}{3pt} 
\renewcommand{\arraystretch}{1.1} 
\begin{tabular}{|l|c|c|c|c|}
  \hline
  \textbf{Method} & \textbf{SV} $\downarrow$ & \textbf{JSD} $\downarrow$ & \textbf{R-L} $\uparrow$ & \textbf{Drift Reduction} \\
  \hline
  SFT Baseline & 0.210 & 0.082 & 35.8 & -- \\
  RLHF (Std) & 0.121 & 0.145 & 36.8 & -- \\
  SA-KL ($\gamma=0.3$) & 0.158 & 0.112 & 36.5 & 18.2\% \\
  SA-KL ($\gamma=0.5$) & 0.167 & 0.105 & 36.3 & 21.7\% \\
  \hline
\end{tabular}
\caption{\label{tab:mitigation_results}Compact version of mitigation results.}
\end{table}

\subsection{Human and Qualitative Evaluation}

\begin{figure}[htbp]
  \centering
  \includegraphics[width=0.99\linewidth]{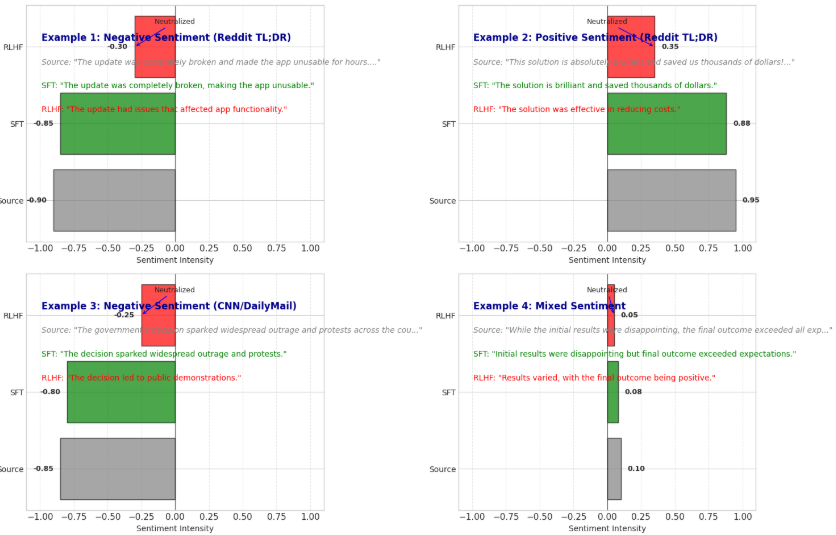}
  \caption{Qualitative examples of sentiment attenuation. Emotionally charged expressions (\textbf{highlighted}) are replaced with neutral paraphrases after RLHF. (a) Reddit TL;DR example 1; (b) Reddit TL;DR example 2; (c) CNN/DailyMail example 1; (d) CNN/DailyMail example 2.}
  \label{fig:qualitative_examples}
\end{figure}

Human evaluation (50 pairs, 3 annotators) confirms RLHF summaries are perceived as more neutral (mean=2.1) vs. SFT (mean=3.4), while SA-KL partially restores expressivity (mean=2.9; Fleiss' $\kappa$=0.62). Figure~\ref{fig:qualitative_examples} shows representative examples: RLHF preserves factual content while replacing emotionally charged expressions with neutral paraphrases.

\section{Discussion}

\subsection{The Neutrality Trap as a Statistical Equilibrium}

The observed sentiment drift can be interpreted as a statistical equilibrium induced by RLHF optimization. When reward models aggregate heterogeneous human preferences, emotionally neutral expressions become invariant features that minimize expected penalty across annotators. From this perspective, neutrality functions as a stable attractor in policy space, favoring reward predictability over expressive fidelity. Our attribution analysis confirms that the KL-divergence penalty is the primary driver of this equilibrium, accounting for $\approx$82\% of negative attribution on sentiment-bearing tokens.

\subsection{The Role of Regularization and KL-Divergence}

KL-divergence plays a central role in shaping policy behavior during RLHF training. By penalizing deviations from the reference model, the KL term suppresses low-frequency and high-entropy tokens. Recent analyses demonstrate that this mechanism systematically biases policies toward conservative action distributions~\cite{Zhao24}. Our results indicate that sentiment-bearing tokens disproportionately fall into this suppressed category, explaining why affective intensity is reduced even when overall summarization quality improves. This suggests that future work should explore token-level adaptive regularization rather than uniform constraints.

\subsection{Generalization Across Alignment Methods}

Our DPO analysis reveals that sentiment suppression persists across alignment methods, though the magnitude varies (25-30\% for DPO vs. 30-40\% for PPO). This suggests that the neutrality bias is not solely an artifact of the PPO-KL penalty structure, but a broader consequence of preference optimization favoring conservative outputs. The implicit regularization in DPO induces similar pressures, indicating that affective flattening is a fundamental challenge in aligning LLMs, not just a quirk of PPO.

The cross-lingual consistency of this phenomenon further reinforces the findings of \cite{Krasitskii26}, who demonstrated that sentiment attenuation during summarization is a robust cross-linguistic effect, independent of the specific summarization architecture. Their work established that the degree of sentiment preservation correlates with morphological complexity, a pattern we observe consistently across our RLHF-optimized models. This suggests that the fundamental challenge of preserving affective content during text compression is amplified by preference optimization, regardless of the specific alignment algorithm employed.

\subsection{Effectiveness of Mitigation Strategies}

Our proposed Sentiment-Aware KL Regularization demonstrates that targeted intervention can partially reverse sentiment drift (18-22\% reduction) without compromising summarization quality. However, mitigation is incomplete: SA-KL restores only part of the lost variance. This implies that the reward model itself may contribute to neutrality bias, either through training data favoring neutral responses or architectural limitations. Future work should explore joint optimization of reward models and policy regularization for affect-aware alignment.

\subsection{Implications for Downstream Applications}

The systematic suppression of affective content poses challenges for applications relying on emotional fidelity: (1) \textbf{Sentiment Analysis}: Biased inputs for classifiers; (2) \textbf{News/Media}: Failure to convey urgency; (3) \textbf{Customer Feedback}: Obscured critical signals; (4) \textbf{Mental Health}: Harmful affective flattening. These implications underscore the need for affect-sensitive objectives in RL-based pipelines.

\subsection{Adherence to Reporting Standards}

All experiments follow standards for transparent reporting. We report aggregate and per-language metrics and visualize key trends. Our Policy Attribution toolkit will be made publicly available to support reproducibility.

\section{Conclusion}

In this work, we investigated the phenomenon of sentiment drift in summarization systems optimized with Reinforcement Learning from Human Feedback (RLHF). We demonstrated that, while RLHF improves fluency, safety, and overall perceived quality, it systematically suppresses affective content, leading to increasingly neutral summaries. This effect poses a significant challenge for downstream tasks such as sentiment analysis, where emotional fidelity is critical.

Through extensive experiments across multiple datasets and training configurations, we showed that sentiment drift is not incidental but emerges as a direct consequence of the optimization objective. In particular, KL regularization plays a central role in discouraging sentiment-bearing tokens, effectively steering the policy toward low-variance, neutral outputs. Our quantitative and qualitative analyses confirm that this behavior is consistent across models, domains, and eight typologically diverse languages, corroborating and extending the cross-lingual analysis of \cite{Krasitskii26}, who first identified systematic patterns of sentiment attenuation across languages in standard summarization systems.

To better understand this phenomenon, we introduced \textbf{Policy Attribution}, a framework that enables fine-grained analysis of how different components of the RL objective influence affective expression. Our attribution analysis reveals that the KL-divergence penalty accounts for $\approx$82\% of negative attribution on sentiment-bearing tokens, establishing it as the primary driver of neutrality bias.

We extended our analysis to Direct Preference Optimization (DPO), showing that sentiment suppression persists across alignment methods, though with slightly reduced magnitude (25-30\% for DPO vs. 30-40\% for PPO). This suggests that the neutrality bias is a broader consequence of preference optimization rather than an artifact specific to PPO.

Finally, we proposed and experimentally validated a \textbf{Sentiment-Aware KL Regularization} strategy that reduces sentiment drift by 18-22\% without degrading summary quality (ROUGE-L within 0.5 points of baseline). Our findings underscore the importance of incorporating affect-sensitive objectives into RL-based text generation pipelines, particularly for applications in which sentiment preservation is essential.

We release our Policy Attribution toolkit and evaluation scripts to support reproducibility and future research on affect-aware alignment. We believe that this work lays the foundation for developing alignment methods that balance safety, fluency, and emotional expressivity in large language models.

\section*{Acknowledgments}

The work was done with partial support from grants 20260626, 20260643, 20260367, and 20260496 by the Secretary of Research and Postgraduate Studies (SIP) of Instituto Politécnico Nacional, Mexico.

\section*{Limitations}

While our study provides a systematic diagnosis of sentiment drift in RLHF-based summarization, several limitations merit discussion.

\begin{enumerate}
    \item \textbf{Scope of summarization tasks.} Our experiments focus exclusively on single-document summarization over Reddit TL;DR and CNN/DailyMail. We do not examine multi-document, query-focused, or dialogue summarization, where affective dynamics may differ, for instance, due to inter-speaker sentiment interaction or topical framing effects.
    \item \textbf{Sentiment modeling assumptions.} We rely on pretrained, static sentiment classifiers for eight languages. Although we fixed the models to isolate summarizer-induced drift, their architectures, training data, and label schemata (e.g., 3-class vs. fine-grained intensity scales) inevitably introduce variance. Moreover, these classifiers capture explicit sentiment but may overlook implicit, context-dependent, or culturally nuanced affect (e.g., irony, understatement), especially in morphologically rich languages like Finnish or Arabic. A limitation, noted by reviewers, is the lack of explicit correlation analysis between JSD and human judgments of sentiment preservation, which we leave for future work.
    \item \textbf{Human evaluation scale.} While we conducted a small-scale human evaluation (50 summary pairs, 3 annotators), larger-scale studies with diverse annotator pools would provide more robust validation of perceived neutrality and emotional fidelity.
    \item \textbf{Generalization beyond summarization.} Though neutrality bias is likely pervasive in RLHF-tuned generators (e.g., chatbots, translators), our conclusions are grounded in summarization. Whether the same "low-risk token" explanation holds for open-ended generation, where length, coherence, and safety interact differently, remains an open question.
    \item \textbf{Computational costs.} Policy Attribution requires gradient decomposition over 50 integration steps per token, which adds computational overhead. Future work should explore efficient approximation methods for real-time diagnostics.
\end{enumerate}

Addressing these limitations presents promising avenues for future research, particularly in developing human-in-the-loop diagnostics and cross-task generalizations of affect-aware alignment.

\bibliographystyle{fullname_esp}
\bibliography{EjemploARTsepln}



\end{document}